\documentclass[smallabstract,smallcaptions]{dccpaper}

\usepackage{epsfig}
\usepackage{citesort}
\usepackage{amsmath}
\usepackage{amssymb}
\usepackage{color}
\usepackage{url}

\usepackage{algorithm}
\usepackage{algorithmic}
\usepackage{amsfonts}       
\usepackage{nicefrac}       
\usepackage{microtype}      
\usepackage{amsthm}
{
   \theoremstyle{plain}

}
\usepackage{booktabs}       
\usepackage{authblk}

\newlength{\figurewidth}
\newlength{\smallfigurewidth}

\begin{document}

\title
{\large
\textbf{DNQ: Dynamic Network Quantization}
}

\author[1]{Yuhui~Xu}
\author[2]{Shuai~Zhang}
\author[2]{Yingyong~Qi}
\author[1]{Jiaxian~Guo}
\author[1]{Weiyao~Lin}
\author[1]{Hongkai~Xiong}
\affil[1]{\footnotesize Department of Electronic Engineering, 
	Shanghai Jiao Tong University, 
	Email: (yuhuixu, gjx925825293, wylin, xionghongkai)@sjtu.edu.cn}
\affil[2]{\footnotesize Qualcomm AI Research,
	Email: (shuazhan, yingyong)@qti.qualcomm.com}
\renewcommand\Authands{ and }

\maketitle
\thispagestyle{empty}

\begin{abstract}
  Network quantization is an effective method for the deployment of neural networks on memory and energy constrained mobile devices. In this paper, we propose a Dynamic Network Quantization (DNQ) framework which is composed of two modules: a bit-width controller and a quantizer. Unlike most existing quantization methods that use a universal quantization bit-width for the whole network, we utilize policy gradient to train an agent to learn the bit-width of each layer by the bit-width controller. This controller can make a trade-off between accuracy and compression ratio. Given the quantization bit-width sequence, the quantizer adopts the quantization distance as the criterion of the weights importance during quantization. We extensively validate the proposed approach on various main-stream neural networks and obtain impressive results.
\end{abstract}

\section{Introduction}
Deep convolutional neural networks have shown their strengths in computer vision and machine learning. In order to achieve better performance, deeper and wider networks are designed which poses heavy burden on storage and computational resources. It is challenging to deploy neural networks on memory and energy constrained devices such as mobile phones and drones. To solve this problem, low precision networks are proposed and attract many researchers interests. Quantized networks with low-precision weights can achieve competitive performance with the full-precision models. Network quantization can not only reduce the memory size, but also reduce the computation cost when acceleration engines \cite{han2016eie} are used.

Most of the existing network quantization works, the quantization bit-widths are usually kept the same for different layers \cite{li2016ternary} \cite{rastegari2016xnor}  \cite{zhou2017incremental}. However, since the representation abilities and capacities of layers are different, we address that the bit-width should be learned to conform to its representation ability. Moreover, after the bit-width is determined, it is also important to quantize the network and preserve its accuracy at the same time. Recently proposed iterative quantization method \cite{zhou2017incremental} \cite{xu2018Quantization} that incrementally partitions the weights into a quantization part and re-training part achieves good performance. However, INQ \cite{zhou2017incremental} determines the importance of network weights by simply comparing their absolute values, which is less convincing and may have limitations in guaranteeing the performance after quantization. In this paper, we argue that quantization distance is a better criterion.



The contributions of this paper are:
\begin{itemize} 
    \setlength\itemsep{0em}
    \item[1.] We propose a \textbf{Dynamic Network Quantization (DNQ)} framework which consists of two modules: a bit-width controller and a quantizer. 
    \item[2.] The bit-width in each layer is modeled by a Markov decision process (MDP). We train a policy network by policy gradient \cite{sutton2000policy} to search for the optimal quantization bit-width for each layer, such that each layer can be adaptively quantized by an optimal bit-width. 
    \item[3.] A distance-based criterion is introduced during iteratively network quantization.  
    \item[4.] The experiment results demonstrate that DNQ can preserve a low accuracy drop with large compression ratio.
\end{itemize}

The rest of this paper is organized as follows. Some related works are summarized in section \ref{prior_work}. In section \ref{DNQ}, we describe Dynamic Network Quantization (DNQ) framework. The experimental results are reported in section \ref{experiments}, followed by the conclusion in Section \ref{conclusion}.

\section{Related Work} \label{prior_work}
\textbf{Network quantization:} Many researches obtain network compression by network quantization. Basically, network compression aims to group weights with similar values to reduce the number of free parameters. Hash-net \cite{chen2015compressing} constrains weights hashed into different groups before training. Within each group the weights are shared and only the shared weights and hash indices need to be stored. Training binary and ternary networks attracts many researchers interests \cite{zhu2016trained} \cite{shuai-LWB2018} \cite{courbariaux2015binaryconnect}, weights are constrained to be binary or ternary representations during training. Hao Li $et\ al.$ \cite{li2017training} and Yin \cite{shuai-Binaryrelax2018} give a deeper understanding in theory. Different from training low precision networks from scratch, quantizing the pre-trained full-precision model can better preserve the accuracy of the network. Han $et\ al.$\cite{han2015deep} present deep compression which combines the pruning \cite{han2015learning}, vector quantization and Huffman coding, and reduces the model size by 35$\times$ on AlexNet and 49$\times$ on VGG-16. Iterative quantization \cite{zhou2017incremental} \cite{xu2018Quantization} utilizes the characteristic of the network itself. This method partitions the weights into two different parts: one part is used to quantize and another part is used to retrain to compensate for quantization loss.


\textbf{Reinforcement learning:} Reinforcement learning has been proven an effective tool to solve many tasks. Neural architecture search has shown its strength by searching for the optimal architecture of a given data-set. \cite{zoph2016neural} uses a controller to sample child networks of different architectures. Different from random search, NASnet \cite{zoph2017learning} propose a new search space to achieve better performance. Efficient architecture search algorithms \cite{pham2018efficient} \cite{liu2017hierarchical} are proposed to reduce the time cost and computation cost of architecture search. Deep reinforcement learning has also been applied in network compression. Ji $et\ al.$\cite{lin2017runtime} prune filters of the network based on different input images by Deep Q-learning. He $et\ al.$ \cite{he2018adc} find the sparsity of each layer by reinforcement learning.

\section{Dynamic network quantization} \label{DNQ}
\begin{figure}
  \centering
  \includegraphics[width=4in]{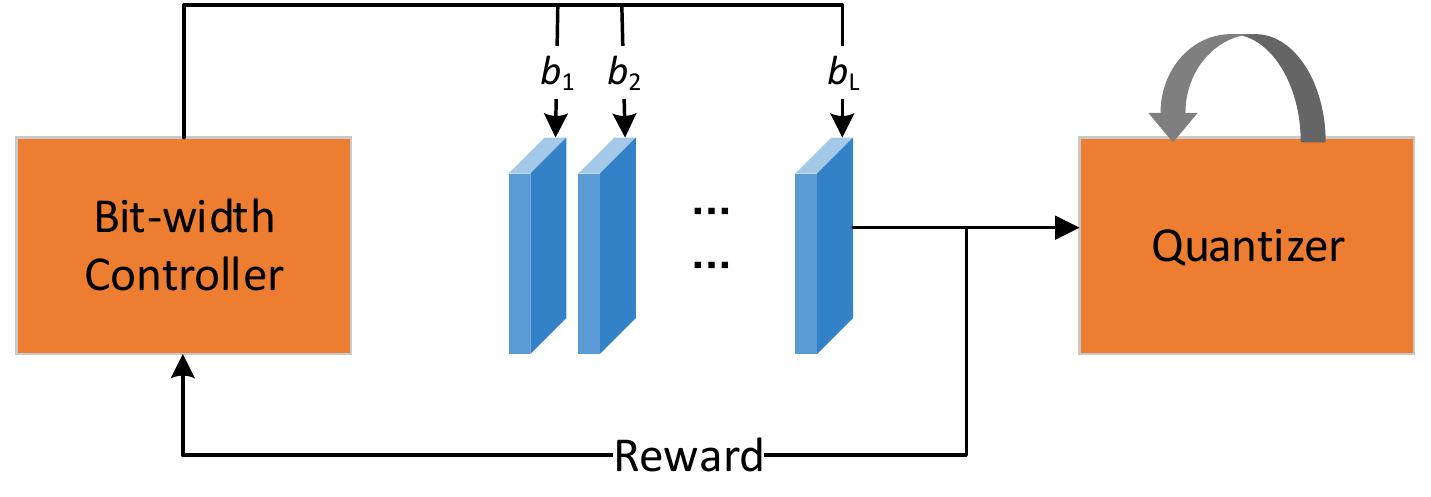}\\
  \caption{The framework of DNQ. DNQ is composed of two modules. The first module is a bit-width controller which training a policy network to generate the bit-width for each layer. The second module is a quantizer that quantizes the network iteratively based on the given bit-width sequence.}\label{fig.1}
\end{figure}

The framework of our Dynamic Network Quantization (DNQ) is consisted of two modules (see Figure~\ref{fig.1}):
\begin{itemize}
    \setlength\itemsep{0em}
    \item \textbf{Module 1}: a bit-width controller that generates optimal quantization bit-width for each layer;
    \item \textbf{Module 2}: a quantizer which iteratively quantizes the weights according to the bit-width given by the bit-width controller.
\end{itemize}


\subsection{Bit-width controller -- module 1}
The compression ratio is related with the quantization bit-width $b_l$  which indicates that $b_l$ bits are used to encode the index of quantized weights. If we use $B$ bits to store one floating point weight, the compression ratio can be expressed as follows:
\begin{equation}\label{equ.1}
\begin{split}
  & r=\frac{\sum_{l=1}^L n_l B}{\sum_{l=1}^L n_lb_l+k_lB}\\
\end{split}
\end{equation}

Our bit-width controller in this work is based on reinforcement learning for training an agent to maximize the cumulative reward when interacting with an environment. This problem is solved by training a policy network $M_\theta$, the input sequence is the embedding of the network and the output sequence $B_L=(b_1,\ldots,b_l,\ldots,b_L)$ is the bit-widths of the network, where $b_l$ is the bit-width of the $l^{th}$ layer. In time step $l$, the state $s$ is the current produced bit-width sequence $(b_1,\ldots,b_{l-1},)$. The action ${a_l}$ we choose in time step $l$ indicates the bit-width used to quantize the layer, where $a_l\in (2,3,\ldots,8)$.
%
The bit-width sequence directly affects the accuracy of the network and it is meaningless to provide a model with high compression rate and low accuracy. There is a tradeoff between the compression ratio and the performance. Thus, the reward $R$ is defined as:
\begin{equation}\label{equ.2}
\begin{split}
  & R=Acc+\lambda r, \\
\end{split}
\end{equation}
where $Acc$ is the accuracy of the quantized network without fine-tuning and $r$ is the compression ratio. $\lambda$ is a positive real value.

$R$ is the reward of the finished sequence $(b_1,\ldots,b_l,\ldots,b_L)$. We should not only consider the fitness of previous layers' bit-widths but also the future outcome. Therefore, to evaluate the action $a_t$ in time step t, we apply Monte Carlo search to sample the next $L-t$ bit-widths. We average the $N$ times sampling results to reduce the variance:
\begin{equation}\label{equ.3}
\begin{split}
  & R^{M_\theta}(s_t=B_{t-1},a_t=b_t)=\frac{1}{N}\sum_{n=1}^N R_n(B_L),\quad B_L= MC(B_t;N), \\
\end{split}
\end{equation}
where $MC(:)$ is the Monte Carlo sampling function.

We train our policy networks by policy gradient \cite{sutton2000policy}:
\begin{equation}\label{equ.4}
\begin{split}
  \nabla_\theta J(\theta)&=\sum_{l=1}^LE_{b_{1:L}~P_\theta}[\nabla_\theta log P_\theta(b_l|b_{1:l-1})R^{M_\theta}]\\
  &=\frac{1}{N}\sum_{n=1}^N \sum_{l=1}^LE_{b_{1:L}~P_\theta}[\nabla_\theta log P_\theta(b_l|b_{1:l-1})R_n(B_L)]
\end{split}
\end{equation}
where $N$ is the Monte Carlo sampling times. $L$ is the total length of the sequence. $P_\theta(b_l|b_{1:l-1})$ is the action probability of action $b_l$ in the time step $l$ given previous $l-1$ actions $b_{1:l-1}$. Algorithm \ref{alg1} details the training procedure of the bit-width controller.

\begin{figure}
  \centering
  \includegraphics[width=4in]{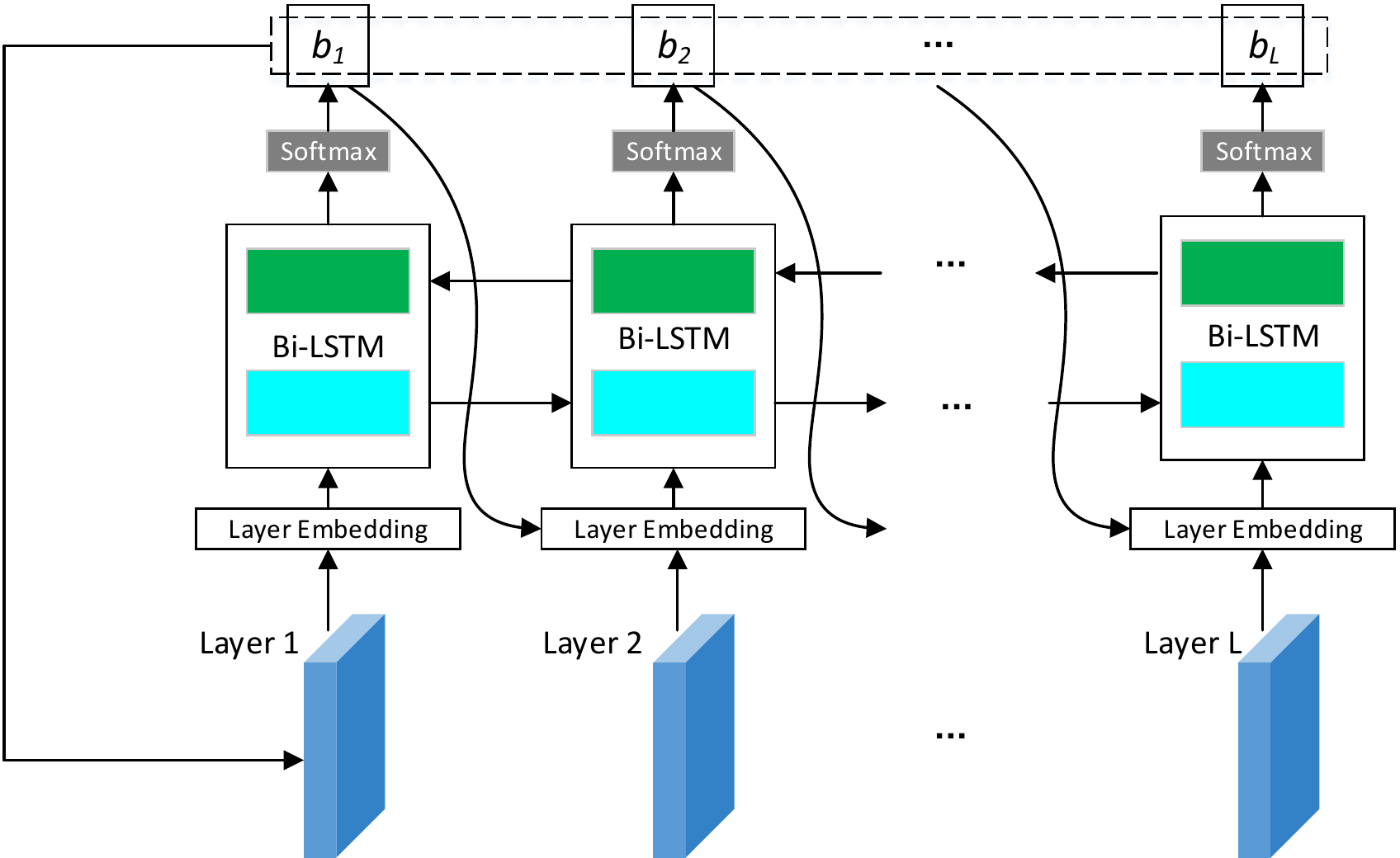}\\
  \caption{The network structure of the bit-width controller. The policy network is a BiLSTM. The input sequence is the embedding of each layer while the output sequence is the nit-width of the layer.}\label{fig.2}
\end{figure}
{
\begin{algorithm}[t]
\renewcommand{\baselinestretch}{1.0}
\renewcommand{\arraystretch}{1.0}
\small
\caption{Bite-width Controller}\label{alg1}
\begin{algorithmic}[1]
\STATE\textbf{Input:}  The pre-trained full-precision DNN model\\
\STATE\textbf{output:} The optimal quantization bit-width sequence for the model\\
\FOR {Iterations}
\STATE Input the embedding sequence of the full precision network by policy network $M_\theta$.
\STATE Generate bit-width sequence $B_L(b_1,\ldots,b_L)$
\FOR {$t$ in $1:L$}
\STATE Compute $R^{M_\theta}(s_t=B_{t-1},a_t=b_t)$ by Eq.\ref{equ.3}
\ENDFOR
\STATE Update policy network $M_\theta$ by policy gradient by Eq.\ref{equ.4}.
\ENDFOR
\end{algorithmic}
\end{algorithm}
}

\subsection{Quantizer -- module 2}
\begin{figure}
  \centering
  \includegraphics[width=4in]{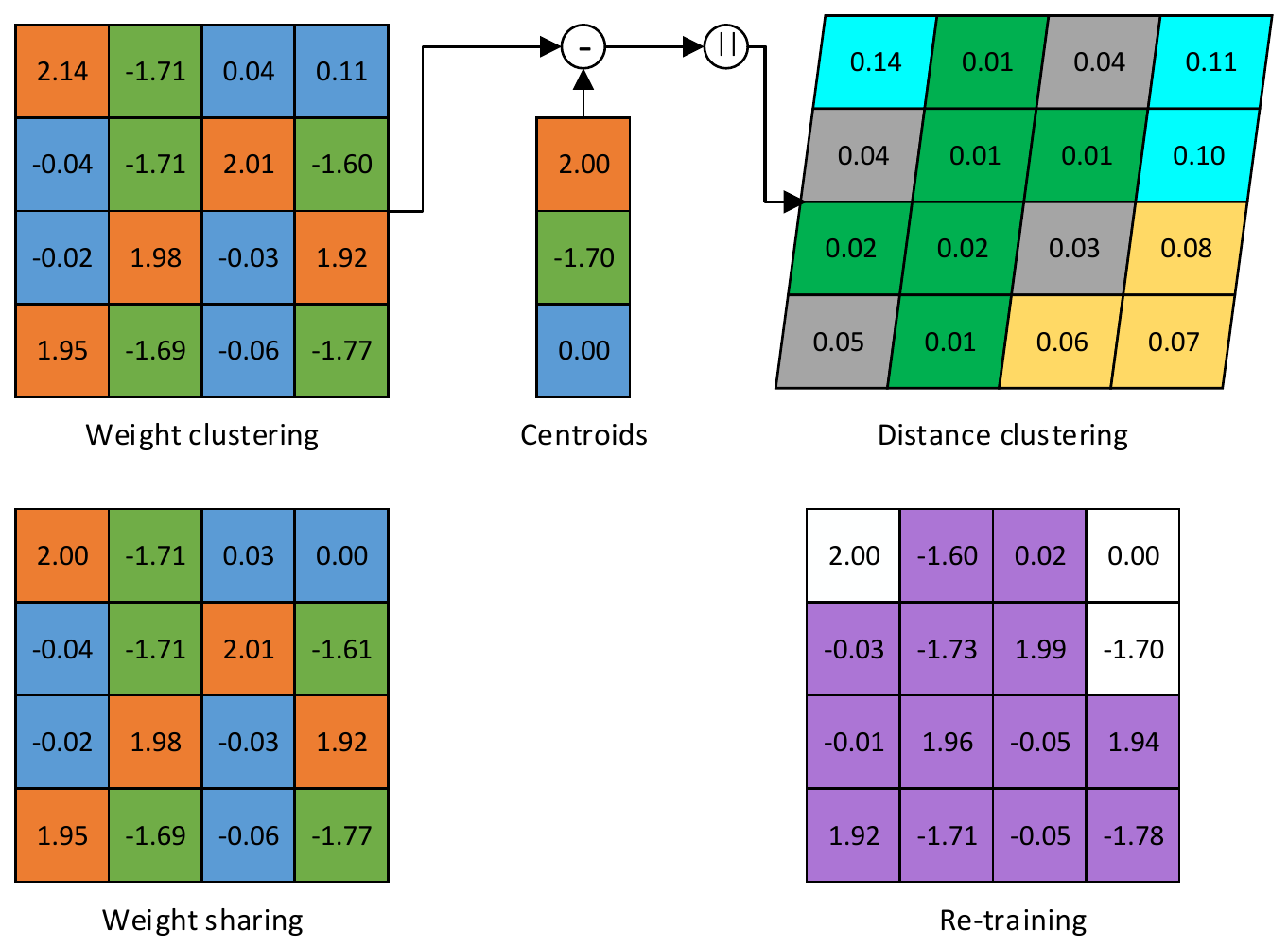}\\
  \caption{Illustration of one iteration of the quantizer. Different color indicates different clusters. Rectangle represents the weight matrix while parallelogram represent the distance matrix. First, the weights of the the network are clustered; Second, the distance between the weights and centroids generated by weight clustering are computed and weights are re-clustered based on distance; Third, weights with bigger distance are quantized in preference; Fourth, accuracy is recovered by re-training. }\label{fig.3}
\end{figure}

As the quantization bit-width is determined, the quantizer maps the weights to their corresponding centroids and recovers the accuracy of the quantized network. Inspired by the INQ \cite{zhou2017incremental}, we follow an iterative strategy that quantizes part of the weights and retrains the remaining weights to recover accuracy. Different from INQ , we use k-means to control the quantization loss and we use quantization distance as the criterion of the weights importance. The quantization part can be divided into four steps:\emph{ weight clustering, distance clustering, weight-sharing and re-training} (as shown in Fg.\ref{fig.3}).

\textbf{Weight clustering:} We adopt the commonly used k-means clustering. By minimizing the within-cluster sum of squares,
\begin{equation}\label{equ.5}
\begin{split}
  & \arg \min  \limits_{C_{1},C_{2}\ldots C_{k}}\sum_{i=1}^{k}\sum_{\omega\in C_{i}}|\omega-c_{i}|^{2}, \\
\end{split}
\end{equation}
where $C_{i}$ is the $i^{th}$ cluster, $c_{i}$ is its corresponding centroid which equals to the mean of the weights in $C_{i}$. Our quantization algorithm is an iterative one that each iteration will contain the clustering step. Note that we only update the centroid in the weight clustering of the first iteration. In other iterations, we fix the centroid and re-allocate the weights into different clusters.

\textbf{Quantization distance clustering:} Quantizing weight $\omega$ to its corresponding centroid $\hat{\omega}$ ($\hat{\omega} \in \{c_{i}\}$) will cause accuracy loss. The bigger the accuracy loss is, the harder the accuracy can be recovered.
\begin{equation}\label{equ.6}
\begin{split}
d=|\omega-\hat{\omega}|
\end{split}
\end{equation}
The quantization loss is highly related with the item $d$ (Eq.\ref{equ.6}). If Loss $L$ is smooth,
\begin{equation}\label{equ.7}
\begin{split}
  &|L(\chi,\omega)-L(\chi,\hat{\omega})|
  = O(|\omega-\hat{\omega}|). \\
\end{split}
\end{equation}

 We can assume that there is a positive correlation between the quantization loss and quantization distance $d$. Based on the observation, the quantization distance $d$ is better used to measure the importance of weights. It can also be assumed that weights with similar quantization distance would show similar effect to the network. We represent each weight of the network by a triplet $\Omega=(\omega,\hat{\omega},d)$. This triplet contains three most important elements during the quantization of $\Omega$. We propose quantization distance clustering to cluster the weights with similar quantization distance. Weights in the same distance cluster will be quantized at the same time.

\textbf{Weight sharing and re-training:} Weights with similar quantization distance are clustered together. The weights in the large distance cluster are quantized in preference since these weights are more important and need more weights to be re-trained to recover accuracy according to our assumption. Also, the number of weights quantized in the iterations is constrained to be a descending sequence.

A mask matrix $M(p,q)$, which has the same size as weight matrix $W(p,q)$, acts as an indicator function to fix the quantized weights:
\begin{equation}\label{equ.8}
M(p,q)=\left\{
\begin{aligned}
0 & \quad , \text{if}\ \omega \text{\ is \ quantized}\\
1 &  \quad , \text{otherwise}
\end{aligned}
\right.
\end{equation}
We re-train the network using stochastic gradient decent(SGD) to update the un-quantized weights. To fix the quantized weights, we use the indicator function $M$ as a mask on the gradient of the weights to control the gradient propagation:
\begin{equation}\label{equ.9}
\begin{split}
  \omega\leftarrow \omega-\gamma \frac{\partial E}{\partial (\omega)}M
\end{split}
\end{equation}
Algorithm \ref{alg2} details the iterative quantization procedure.

{
\begin{algorithm}[t]
\renewcommand{\baselinestretch}{1.0}
\renewcommand{\arraystretch}{1.0}
\small
\caption{Quantizer}\label{alg2}
\begin{algorithmic}[1]
\STATE\textbf{Input:}  Bit-width sequence and pre-trained full-precision DNN model\\
\STATE\textbf{output:} The quantized low-precision model\\
\FOR {Iterations}
\STATE Given bit-width $B_L$, conduct weight clustering as Eq.\ref{equ.5}.
\STATE Conduct distance clustering
\STATE Quantize the weights in the bigger distance cluster.
\STATE Re-train the un-quantized weights to recover accuracy.
\ENDFOR
\STATE All weights are quantized.
\end{algorithmic}
\end{algorithm}
}

\section{Experiments} \label{experiments}
We conduct experiments on two data-sets CIFAR-10 and ILSVRC2012 (ImageNet-12) to evaluate our proposed framework DNQ. CIFAR-10 consists of 60,000 RGB images (32$\times$32) with 10 classes, with 6,000 images per class. There are 50,000 training images and 10,000 test images.
ILSVRC2012 contains as much as 1,000 classes of objects with nearly 1.2 million training images and 50 thousand validation images. For CIFAR-10, we use a 5-layer network CIFAR-Net \footnote{CIFAR-NET model file can be found: https://github.com/BVLC/caffe/tree/master/examples/cifar10}. For ILSVRC2012, we use AlexNet and ResNet18 for evaluation. We compare our DNQ with state-of-the-art quantization methods on ILSVRC2012.

\subsection{Implementation Details}
The implementation of DNQ is consisted of two modules. The first is the bit-width controller and the second is a quantizer. The module 1 generates bit-width for each convolutional layer and module 2 conducts quantization according to the bit-width. The bit-width controller is implemented on Tensorflow  while the quantizer is implemented on Caffe.

In the module 1, we use a bidirectional LSTM as the policy network (as shown in Fg.\ref{fig.2}). The policy network is trained using SGD with learning rate 0.01. The batch size is 5 and the total iterations are 1,000. In the bit-width controller, we only consider the convolutional layers of the network for the following reasons. First, in a network, convolutional layers extracting features are more important and more difficult to be quantized than fully-connected (FC) layers. Second, most of the computation cost is caused by convolution operation. The computation cost can be reduced if the bit-width is smaller when using the acceleration engine \cite{han2016eie}. Third, recent networks like ResNet \cite{he2016deep} use global average pooling (GAP) other than redundant FC layers. The bit-width of the FC layers in CIFAR-Net and AlexNet is fixed to 3.

In the module 2, the weight clustering cluster number k equals $2^{b-1}+1$ because we use one bit to store zero. The distance clustering cluster number is 12. CIFAR-Net, AlexNet and ResNet are all trained using SGD. In each iteration, CIFAR-Net is retrained with learning rate 0.01 for total 8k iterations, with batch size 100. AlexNet and ResNet are both retrained with an initial learning rate 0.001, decaying by a factor of 10 after 30k, 60k iterations for total 60k iterations. The batch size of AlexNet is 256 and the batch size of ResNet is 80. For extremely low bit-with, we do the distance clustering within each weight cluster.
\begin{figure}
  \centering
  \includegraphics[width=5in]{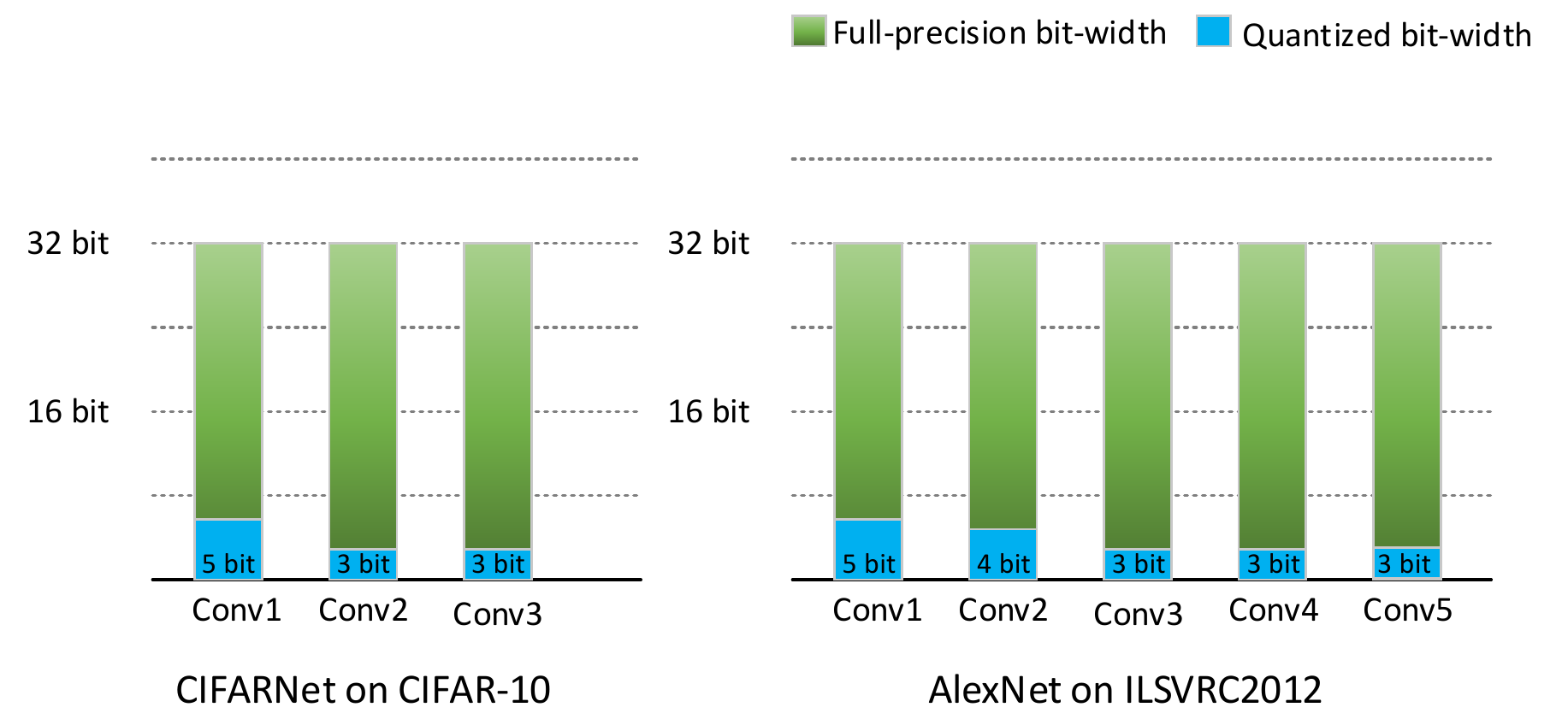}\\
  \caption{The bit-width generated by the bit-width controller. The left part is the bit-width of each convolutional layers of CIFAR-Net while the right part is the bit-width of each convolutional layers of AlexNet}\label{fig.4}
\end{figure}

\subsection{Results on CIFAR-10}
\begin{table}
	\caption{Experiments of DNQ on CIFAR-Net (module 1 and module 2 indicate the bit-width controller and the quantizer resectively)}
	\label{table1}
	\centering
	\begin{tabular}{lllll}
		\toprule
		Method & Bit-width  & Accuracy(\%)    & Compression ratio (Eq.\ref{equ.1}) \\
		\midrule
		INQ \cite{zhou2017incremental}& 5 & 81.08 &6.3 \\
		SLQ \cite{xu2018Quantization} & 5 & 81.11 &6.3 \\
		DNQ ((module 2)               & 5 & 81.63 &6.3 \\
		\midrule
		INQ                           & 3 &79.52&10.6 \\
		DNQ (module 2)                & 3 &79.89&10.6 \\
		DNQ(module 1+ 2)       & Dynamic &80.56 &10.5 \\
		\bottomrule
	\end{tabular}
\end{table}

For CIFAR-10, we use a 5 layer convolutional network with 3 convolutional layers. We first test the effectiveness of our quantizer which is the module 2 in DNQ. The result is shown in Table.\ref{table1}. All layers are quantized into 5 bit or 3 bit when only using the module 2. The accuracies of the 5-bit and 3-bit DNQ quantized CIFAR-Net are higher than the INQ 5-bit and 3-bit quantized models respectively. Besides, the 5-bit model quantized by the proposed quantizer also outperforms the 5-bit model quantized by SLQ \cite{xu2018Quantization}. This demonstrates the effectiveness of the quantization distance criterion. For module 1, the bit-widths of all three convolutional layers are shown in Figure \ref{fig.4}. The bit-width of conv1 is 5 while the other two convolutional layers are both 3. The result of our DNQ applied on CIFAR-Net is as shown in Table.\ref{table1}. With module 1 and 2 both applied, the quantized CIFAR-Net by DNQ can obtain better performance with nearly the same compression ratio compared with other methods INQ with universal bit-width 3. The quantized model by DNQ also achieves better performance the model quantized by only using module 2.

\subsection{Results on ILSVRC2012}
For ILSVRC2012, we use AlexNet and ResNet18 for experiments. The bit-width of ALexNet is shown in Fg.\ref{fig.4}. The result is shown in Table.\ref{table2}. AlexNet quantized by DNQ can achieve negligible accuracy loss compared with full-precision model. Note that we do not add BN layers and we quantize all the layers in the AlexNet. The quantized ResNet18 can achieve nearly the same performance as full-precision model.

To show the effectiveness of our approach, we compare the proposed DNQ with INQ \cite{zhou2017incremental} on ResNet18. The results are shown in Table.\ref{table2}. The bit-width for INQ is set 3 for all of the layers. The bit-width for DNQ is determined by the bit-width controller (module 1). The compression ratio for INQ is 10.7, while the compression ratio for our DNQ is 10.6 which is nearly the same. However, DNQ can have better result than INQ since we use the dynamic bit-widths and distance based quantization.

\begin{table}
  \caption{Experiments of DNQ on ILSVRC2012}
  \label{table2}
  \centering
  \begin{tabular}{l|l|l|l}
    \toprule
    Network & Bit-width  & Top-1 Accuracy (\%) 
    & Compression ratio (Eq.\ref{equ.1}) \\
    \midrule
    AlexNet ref                     & 32          & 56.81          &         \\
    AlexNet-DNQ                     & Dynamic    & 56.72 (-0.09)   & 10.6    \\
    ResNet18 ref                    & 32          & 68.20          &         \\
    ResNet18-INQ \cite{zhou2017incremental}  &3  & 68.08 (-0.12)   & 10.7    \\
    ResNet18-DNQ                    & Dynamic     & 68.23 (+0.03)  & 10.6    \\
    \bottomrule
  \end{tabular}
\end{table}

\section{Conclusion} \label{conclusion}
In this paper, we propose Dynamic Network Quantization (DNQ) framework which is composed of two modules. First, we train an agent network by policy gradient \cite{sutton2000policy} to learning the bit-width for each layer. Second, we propose a quantization distance based quantizer to quantize the network iteratively based on the given bit-width. We evaluate our DNQ with CIFAR-Net on CIFAR-10 and AlexNet, ResNet18 on ILSVRC2012 and achieve impressive results.

\section{References}
\bibliographystyle{IEEEbib}
\bibliography{refs}

\end{document}